\documentclass[letterpaper, 10 pt, conference]{ieeeconf}  

\IEEEoverridecommandlockouts                              

\usepackage[T1]{fontenc}
\usepackage[utf8]{inputenc}
\usepackage[breaklinks=true,
            colorlinks,
            linkcolor = red,
            urlcolor  = purple, 
            citecolor = blue,
            bookmarks = true]{hyperref}  
\usepackage{array}
\usepackage{booktabs}
\usepackage{tabularx}
\usepackage{makecell}
\usepackage{graphicx}
\usepackage{amsmath}
\usepackage{amssymb}
\usepackage{algorithm}
\usepackage{algpseudocode}
\usepackage{multirow}
\usepackage{xcolor}
\usepackage{cite}
\usepackage{utfsym}

\newcommand{\figpanelref}[2]{%
  Fig.~\hyperref[#1]{\ref*{#1}#2}%
}

\title{\LARGE \bf
COACT: Learning Human--Humanoid \underline{CO}llaborative C\underline{A}rrying via \underline{C}ontact-Guided Motion \underline{T}racking
}

\author{Yibo Xia$^{1,2}$, Rui Li$^{1,2}$, Rongsheng Qu$^{1,2}$, Weixin Li$^{1,2}$, Yunhong Wang$^{1,2}$
\thanks{Corresponding authors: Weixin Li and Yunhong Wang.}
\thanks{$^{1,2}$Yibo Xia, Rui Li, RongSheng Qu, Weixin Li, Yunhong Wang are with State Key Laboratory of Software Development Environment, School of Computer Science and Engineering, Beihang University, Beijing, 100191, China, and also with the Zhongguancun Academy, Beijing, 100094, China.
(e-mail: {\tt\small yiboxia, weixinli, yhwang@buaa.edu.cn})}%
}

\IEEEaftertitletext{\vspace{-12pt}
\begin{minipage}{\textwidth}
  \centering
  \includegraphics[width=\linewidth]{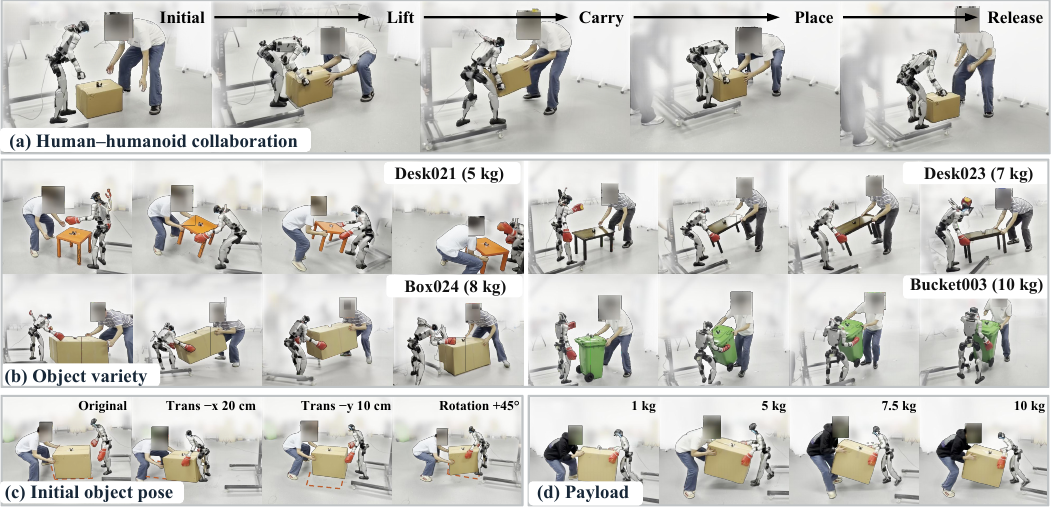}\par
  \nointerlineskip
  \makeatletter
  \def\@captype{figure}
  \setlength{\abovecaptionskip}{3pt}
  \caption{Real-world human--humanoid collaborative carrying:
(a) starting without pre-established robot--object contact, the humanoid autonomously establishes contact, lifts, carries, places, and releases the object;
(b) a single policy across multiple object shapes and masses;
(c) adaptation to varied initial object poses; and
(d) carrying with payloads of $1$, $5$, $7.5$, and $10$~kg. Within each group in (c) and (d), the object and reference motion are fixed.}
  \label{fig:teaser}
  \label{fig:real_world}
  \makeatother
\end{minipage}
}

\begin{document}

\maketitle
\thispagestyle{empty}
\pagestyle{empty}

\begin{abstract}
Human--humanoid collaborative carrying requires establishing contact for lifting, maintaining it during carrying, and releasing it after placing.
Existing approaches either rely on model-based pipelines with hand-crafted subtasks and per-subtask contact constraints, or learn only lifting or carrying with pre-established robot--object contact.
To enable lifting--carrying--placing behavior from an initially unheld object
without hand-crafted subtasks, we propose COACT, a contact-guided motion tracking framework that learns a single policy from human--object--human demonstrations, with demonstrated contact as explicit guidance for both retargeting and policy learning.
We first introduce COACT-Retarget, a contact-guided dynamic retargeting method that converts imperfect human demonstrations into dynamically feasible references with stable hand–object contact and minimal penetration.
Building on these high-quality trajectories, we further propose COACT-Policy, a closed-loop collaborative policy learning procedure that couples COACT-Partner, a physical co-actuating partner, with a single humanoid policy through shared object dynamics.
Contact labels guide partner assistance and humanoid contact behavior, enabling the policy to learn contact transitions, whole-body coordination, load sharing, and adaptation across multiple objects and motions.
In simulation, the policy completes the full lifting–carrying–placing sequence on 83.86\% of the training references and achieves 46.78\% under large initial object-pose perturbations, outperforming policies trained on other retargeting references in both settings.
On real hardware, the same policy achieves 76.67\% success over 120 trials across 7 objects and 12 lifting--carrying--placing sequences.
\end{abstract}

\vspace{-0.1cm}
\section{INTRODUCTION}
\vspace{-0.1cm}

Humanoid motion tracking~\cite{pengDeepMimic2018,liaoBeyondMimicMotionTracking2025,luoSONIC2026}, teleoperation~\cite{zeTWIST22026,liCLONE2025}, and whole-body loco-manipulation~\cite{xieGRAILGeneratingHumanoid2026,wuSUGARScalableHumanvideodriven2026a,zhaoResMimicGeneralMotion2025} have advanced rapidly, demonstrating increasingly versatile and robust control, but progress in physical human--humanoid collaboration remains limited~\cite{duLearningHumanHumanoidCoordination2025,kumbharEfficientCompliantControl2025,bethalaH2COMPACTHumanHumanoidCoManipulation2025}.
Human--humanoid collaborative carrying is a representative task of such collaboration, which matters in domestic assistance, healthcare, and manufacturing~\cite{agravanteHumanHumanoidCollaborativeCarrying2019}.
The task asks the human and humanoid to jointly lift an object, carry it to a target location, and place it down.

Prior model-based approaches~\cite{agravanteHumanHumanoidCollaborativeCarrying2019} decompose collaborative carrying into hand-crafted subtasks with manually specified transitions and constraints, limiting generalization.
Recent learning-based methods largely focus on collaborative lifting or carrying with initial robot--object contact (first five rows of Table~\ref{tab:setting_comparison}), leaving the complete task unaddressed and limiting real-world applicability.
To this end, we study the full lifting--carrying--placing behavior with a single learned policy starting without initial robot--object contact.

Learning such collaborative behavior from scratch is challenging because the policy must jointly learn whole-body coordination, load-bearing physical contact, and adaptation to the partner's dynamics in a high-dimensional, contact-rich setting.
Exploration therefore relies heavily on carefully engineered rewards~\cite{yangOmniRetargetInteractionPreservingData2025}.
Human--object--human interactions, by contrast, naturally exhibit coordinated whole-body motions and contact transitions, motivating us to learn collaboration from human demonstrations via motion tracking.

However, converting these demonstrations into robot-executable behavior presents two challenges. 
First, existing methods~\cite{yangOmniRetargetInteractionPreservingData2025, araujoRetargetingMattersGeneral2025} largely formulate retargeting as a kinematic optimization problem, often producing implausible motions with artifacts such as collisions, penetrations, and unnatural object--robot distances in contact-rich collaboration scenarios. 
Second, a pose reference does not specify the physical interaction force required under real-world dynamics, where the humanoid must establish contact with the object while keeping balance and coordinating with its human partner.
Both challenges center on contact: retargeting must recover dynamically feasible hand--object contact, while policy learning must realize and coordinate that contact during interaction.

To address these challenges, we present \textbf{COACT} (\textbf{CO}llaborative \textbf{CA}rrying via \textbf{C}ontact-guided \textbf{T}racking), a contact-guided framework for human--humanoid collaborative carrying where contact provides explicit guidance for both motion retargeting and collaboration-policy learning (Fig.~\ref{fig:teaser}).
First, to overcome the limitations of kinematic retargeting, we propose \textbf{COACT-Retarget}, a contact-guided dynamic retargeting method with a \emph{multi-scale contact reward} for stable hand--object contact and \emph{part-wise penetration constraints} to limit body--object penetration.
The refined references provide high-quality training data and dense learning signals for downstream collaboration-policy learning with reinforcement learning (RL).
Second, we introduce \textbf{COACT-Policy}, a closed-loop collaborative policy learning method by coupling a physical co-actuating partner (\textbf{COACT-Partner}) with the humanoid policy through shared object dynamics.
Demonstrated contact labels determine when partner assistance starts and stops, while rewards guide the policy to realize the corresponding humanoid contacts. 
This contact guidance aligns collaboration transitions across lifting, carrying, and placing, while closed-loop interaction enables the policy to learn load sharing and adaptation to the partner's dynamics within demonstrated references.

Unlike prior model-based methods, our approach learns a single closed-loop collaboration policy without requiring hand-crafted subtasks.
In contrast to learning-based methods limited to lifting or carrying with pre-established robot--object contact, we enable a single humanoid policy to complete the lifting--carrying--placing behavior, autonomously establishing, maintaining, and releasing contact.
In simulation, the policy achieves an 83.86\% collaborative carrying success rate on the training references and 46.78\% under large initial pose perturbations, outperforming policies trained on alternative retargeting references under the same learning setup.
Real-world experiments on the Unitree G1 achieve a $76.67\%$ carrying success rate across 7 objects and 12 motion sequences with a single policy.

\begin{table}[t]
  \centering
  \caption{Comparison of existing Human--humanoid collaborative carrying methods.
  ``w/o Initial contact'' indicates whether the full lifting--carrying--placing task can be completed without initial robot--object contact.}
  \vspace{-0.2cm}
  \label{tab:setting_comparison}
  \fontsize{7.5}{8.4}\selectfont
  \renewcommand{\arraystretch}{1.0}
  \setlength{\tabcolsep}{2pt}
  \setlength{\aboverulesep}{1.4pt}
  \setlength{\belowrulesep}{1.4pt}
  \begin{tabular*}{\columnwidth}{@{\extracolsep{\fill}}lcll@{}}
    \toprule
    \textbf{Method} & \textbf{w/o Initial contact} & \textbf{Task} & \textbf{Policy} \\
    \midrule
    Kumbhar et al.~\cite{kumbharEfficientCompliantControl2025} & \usym{2717} & Carry & Model-based \\
    H2-COMPACT~\cite{bethalaH2COMPACTHumanHumanoidCoManipulation2025} & \usym{2717} & Carry & RL \\
    HALO~\cite{zhangHALOLearningHumanrobot2026} & \usym{2717} & Carry & Multi-agent RL \\
    IO-WBC~\cite{zhangInteractionawareWholebodyControl2026a} & \usym{2717} & Lift / push & Multi-agent RL \\
    COLA~\cite{duLearningHumanHumanoidCoordination2025} & \usym{2717} & Lift / carry & RL from scratch \\
    Agravante et al.~\cite{agravanteHumanHumanoidCollaborativeCarrying2019} & \usym{2713} & Lift--carry--place & Model-based \\
    \midrule
    \textbf{COACT (Ours)} & \usym{2713} & \textbf{Lift--carry--place} & \textbf{RL from Demo} \\
    \bottomrule
  \end{tabular*}
\vspace{-0.7cm}
\end{table}

Our main contributions can be summarized as follows:
\begin{itemize}

\item \textbf{A contact-guided collaborative motion tracking framework.}
We propose COACT to learn full lifting, carrying, and placing collaborative behaviors, without initial robot--object contact and hand-crafted subtasks.

\item \textbf{A contact-guided dynamic retargeting method.}

We develop \textbf{COACT-Retarget} to refine noisy kinematic demonstrations with multi-scale contact rewards and part-wise penetration constraints, recovering hand--object contact while limiting undesired penetration.

\item \textbf{A closed-loop collaborative policy learning procedure.}
We introduce \textbf{COACT-Policy} that couples a physical co-actuating partner with the single policy through shared object dynamics and contact guidance across multiple objects and motions.

\end{itemize}

\vspace{-0.15cm}
\section{RELATED WORK}

\vspace{-0.1cm}
\subsection{Human--Humanoid Collaborative Carrying}

Model-based collaborative carrying regulates physical interaction through predefined subtasks.
Agravante et al.~\cite{agravanteHumanHumanoidCollaborativeCarrying2019} use model predictive control and quadratic-programming whole-body control to execute predefined lifting, carrying, and placing subtasks with specified grasp points.
Other model-based transport methods address force sensing and contact stability with force observers, simplified dynamic models, and load-aware squeeze-force control~\cite{rahemHumanHumanoidRobotCooperative2022,kumbharEfficientCompliantControl2025,carabisSlipAvoidanceDualArm2018}.
These methods require predefined phases, transitions, and contact or postural constraints.

Learning-based methods reduce this design burden and adapt to partner motion~\cite{bethalaH2COMPACTHumanHumanoidCoManipulation2025}.
Some methods use multi-agent RL for coordination and compliance in individual tasks such as carrying~\cite{zhangHALOLearningHumanrobot2026}, lifting, or pushing~\cite{zhangInteractionawareWholebodyControl2026a}. They model the partner as a learning agent and assume pre-established robot--object contact.
COLA~\cite{duLearningHumanHumanoidCoordination2025} learns lifting or carrying with a PD-controlled support body as its partner in a closed-loop object-interaction simulator.
The object starts in the robot's hands, whose joints are fixed in a predefined grasp pose.
These methods address individual tasks with existing contact (Table~\ref{tab:setting_comparison}), leaving the full lifting--carrying--placing sequence with autonomous contact establishment and release outside their scope.
COACT instead learns this full task from human--object--human demonstrations with a single policy, without initial robot--object contact or hand-crafted subtasks.

\begin{figure*}[t]
  \centering
  \includegraphics[width=\textwidth]{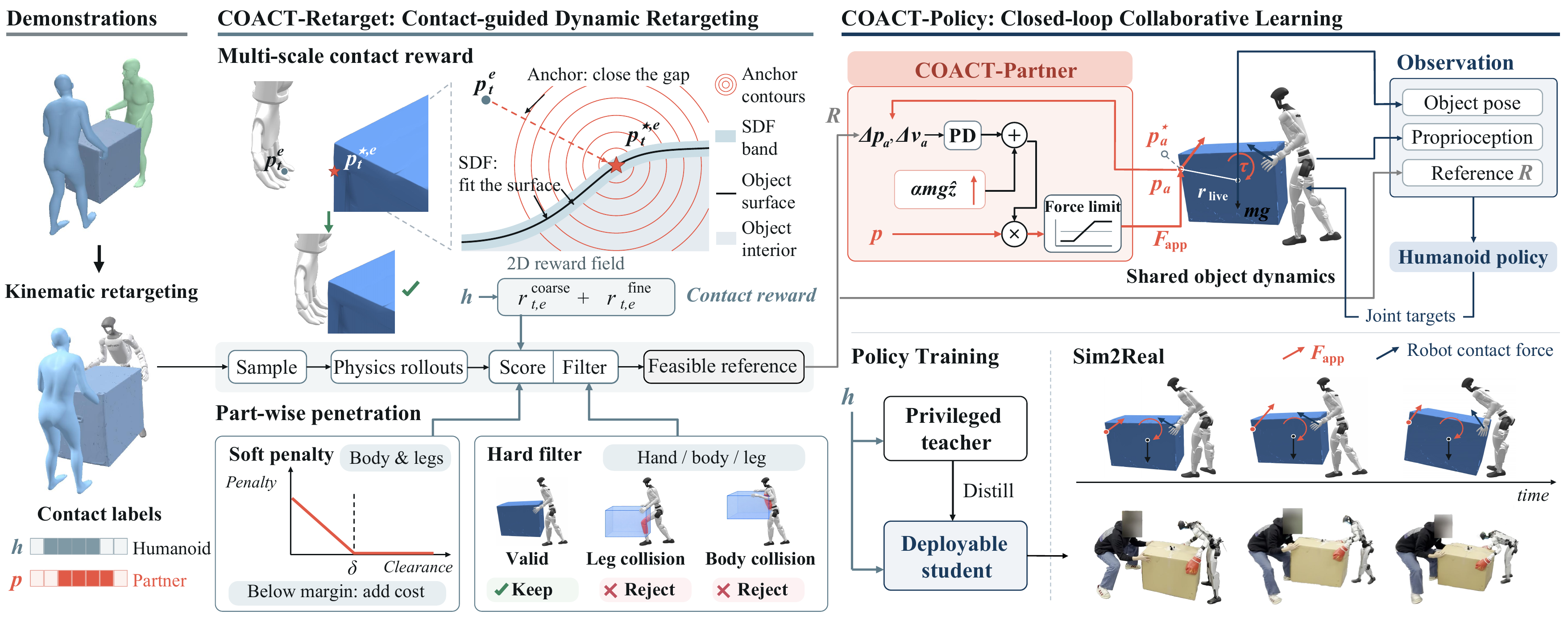}
  \vspace{-0.7cm}
  \caption{\textbf{Overview of the proposed COACT framework.}
  COACT-Retarget refines kinematically retargeted human demonstrations into dynamically feasible references ($R$) using multi-scale contact rewards and part-wise penetration constraints.
  COACT-Policy couples COACT-Partner and the humanoid policy through shared object dynamics.
  Demonstrated partner labels ($p$) gate assistance, while humanoid labels ($h$) specify contact targets.
  A privileged teacher is distilled into a deployable student.}
  \label{fig:pipeline}
\vspace{-0.75cm}
\end{figure*}

\vspace{-0.2cm}
\subsection{Motion Retargeting for Contact-Rich Interaction}
\vspace{-0.1cm}

Motion retargeting bridges the embodiment gap between human demonstrations and humanoid control policies.
Kinematic retargeting methods such as GMR~\cite{araujoRetargetingMattersGeneral2025} mitigate artifacts such as foot sliding and self-penetration through pose alignment, which alone cannot guarantee demonstrated contacts, object support, or clearance of non-contact body parts.
Recent interaction-aware methods~\cite{yangOmniRetargetInteractionPreservingData2025,wuTopoRetargetInteractionpreservingRetargeting2026,fengMinimalistRetargetingGuidedReinforcement2026} incorporate interaction geometry or topology beyond isolated joint or keypoint errors.
OmniRetarget~\cite{yangOmniRetargetInteractionPreservingData2025} represents the humanoid, terrain, and object with an interaction mesh to preserve their spatial and contact relationships, but remains primarily kinematic without modeling interaction dynamics.
Sampling-based trajectory optimization methods, including DynaRetarget~\cite{dhedinDynaRetargetDynamicallyFeasibleRetargeting2026} and SPIDER~\cite{panSPIDERScalablePhysicsInformed2026}, address this limitation by sampling candidate trajectories in a physics simulator for dynamic feasibility.

\vspace{-0.2cm}
\subsection{Motion Tracking for Humanoid Loco-Manipulation}

Physics-based motion imitation and tracking support versatile whole-body humanoid control~\cite{pengDeepMimic2018,liaoBeyondMimicMotionTracking2025,luoSONIC2026,luoPerpetualHumanoidControl2023,heASAP2025,qiHumanoidGPT2026}.
Recent systems further address loco-manipulation and scene interaction~\cite{xieGRAILGeneratingHumanoid2026,liuOpt2Skill2025,fuDemoHLM2026,heULTRA2026,wangPhysHSIRealWorldGeneralizable2025}.
HDMI~\cite{wengHDMILearningInteractive2025} jointly tracks robot and object states from human videos, ResMimic~\cite{zhaoResMimicGeneralMotion2025} augments a general motion tracker with a residual object-aware policy, and SUGAR~\cite{wuSUGARScalableHumanvideodriven2026a} refines human-video priors before distilling them into a hierarchical controller.
Inspired by their success, COACT learns human--humanoid collaborative carrying by tracking retargeted demonstrations with COACT-Partner.
\vspace{-0.15cm}
\section{METHOD}

\vspace{-0.1cm}
\subsection{Overview and Problem Setting}

We learn human--humanoid collaborative carrying from human demonstrations via motion tracking across lifting, carrying, and placing.
Starting without robot--object contact (neither attached nor supported from below), the humanoid must establish contact before lifting, sustain it while sharing the load with the human partner during carrying, and release the object after placing.
We propose COACT (Fig.~\ref{fig:pipeline}), a contact-guided framework in which contact provides explicit guidance from retargeting through policy learning.
First, OmniRetarget~\cite{yangOmniRetargetInteractionPreservingData2025} kinematically retargets demonstrations from CORE4D~\cite{liu2025core4d} to the Unitree G1, producing robot references with object poses and contact labels.
COACT-Retarget (Sec.~\ref{sec:dynamic_retargeting}) then refines these references into dynamically feasible trajectories through sampling-based
optimization with multi-scale contact rewards and part-wise penetration constraints.
Finally, our COACT-Policy (Sec.~\ref{sec:closed_loop_learning}) couples a physical co-actuating partner with the humanoid policy through shared object dynamics, with contact labels guiding partner assistance and humanoid contact behavior.

\vspace{-0.2cm}
\subsection{COACT-Retarget: Contact-Guided Dynamic Retargeting}
\label{sec:dynamic_retargeting}

Conventional kinematic retargeting typically minimizes joint position or orientation errors, resulting in implausible motions with undesired collisions of non-contact body parts, penetrations, and unnatural object--robot distances in contact-rich carrying.
We address these artifacts with \emph{contact-guided dynamic retargeting} (\textbf{COACT-Retarget}), which refines kinematic references in simulation to improve hand--object contact while reducing penetration.

Given a CORE4D motion-capture sequence, we first retarget it to the $29$-DoF Unitree G1 with OmniRetarget~\cite{yangOmniRetargetInteractionPreservingData2025}, producing a kinematic reference state sequence $x^{\mathrm{ref}}_t = \{ q^{\mathrm{ref}}_t , \dot{q}^{\mathrm{ref}}_{t} \}$ where $q^{\mathrm{ref}}_t = \{q^{\mathrm{ref}}_{t,\mathrm{robot}},\,q^{\mathrm{ref}}_{t,\mathrm{obj}}\}$.
The robot configuration $q^{\mathrm{ref}}_{t,\mathrm{robot}}$ comprises joint positions $q^{\mathrm{ref,joint}}_t\in\mathbb{R}^{n_u}$ ($n_u=29$) and base transformation $T^{\mathrm{ref,base}}_t\in SE(3)$, while $q^{\mathrm{ref}}_{t,\mathrm{obj}}\in SE(3)$ has position $p^{\mathrm{ref}}_{o,t}\in\mathbb{R}^3$ and orientation quaternion $\varrho^{\mathrm{ref}}_{o,t}$.
We derive frame-wise contact labels from the original CORE4D data, thresholding each hand-to-object distance at $3$~cm and filling short gaps: $l^{\mathrm{ref,h}}_t\in\{0,1\}^{E}$ records contact for the $E$ end-effectors of the retargeted person, and $l^{\mathrm{ref,p}}_t$ indicates whether either partner hand touches the object.
Each object mesh is converted into an object-local signed distance field (SDF) $\phi$ on a fixed grid.
For a set $S$ of robot geometric elements, $\phi_S(x_t)=\min_{k\in S}\phi(x_t,k)$ queries the closest element, with $\phi_e(x_t)$ for an individual end-effector $e$.

We refine this kinematic reference with a dynamics-aware optimizer that keeps the robot and object motions close to the reference, encourages the intended hand--object contacts, and maintains clearance between the object and non-contact body parts.
Since the resulting optimization problem is high-dimensional, non-smooth, and non-convex, we adopt a sampling-based trajectory optimization approach to solve it.

\subsubsection{Sampling-Based Dynamic Retargeting}
We optimize robot proportional--derivative (PD) targets $U=u_{0:T-1}$ over receding-horizon windows of length $H$ with the sampling-based optimizer~\cite{panSPIDERScalablePhysicsInformed2026} (Alg~\ref{alg:retarget}).
It uses $N_W$ Gaussian control samples per iteration and an elite fraction $\rho$, with controls interpolated between equally spaced knots.
Rollouts follow $x^j_{t+1}=f(x^j_t,u^j_t,t)$, where $j$ indexes candidates, $f$ is the simulator state-transition function, and $x_t$ is the robot--object state.
The accumulated rollout cost is,
{\small
\vspace{-0.1cm}
\begin{equation}
\begin{aligned}
J={}&\sum_t\left\|q_{t,\mathrm{robot}}\ominus q^{\mathrm{ref}}_{t,\mathrm{robot}}\right\|^2_{Q_q}\\[1pt]
&+\sum_t\Big[\left\|p_{o,t}-p^{\mathrm{ref}}_{o,t}\right\|^2_{Q_p}
 +\left\|\varrho_{o,t}\ominus\varrho^{\mathrm{ref}}_{o,t}\right\|^2_{Q_\varrho}\Big]\\[1pt]
&-\sum_{t,e}\left(r^{\mathrm{coarse}}_{t,e}+r^{\mathrm{fine}}_{t,e}\right)
 +\sum_{t,\,P\in\{\mathrm{body},\mathrm{leg}\}}w_Pc^{\mathrm{pen}}_{t,P}
 +J_z
\end{aligned}
\label{eq:retarget_cost}
\end{equation}
}
\vspace{-0.1cm}
where $J_{\mathrm{track}}$ sums the tracking terms weighted by $Q_q$, $Q_p$, and $Q_\varrho$, $w_P$ weights the clearance penalty of group $P$, $\ominus$ denotes a tangent-space pose difference, and $J_z$ penalizes robot--ground penetration at the torso and feet.

\begingroup
\makeatletter
\let\coact@algorithm@style\fst@algorithm
\def\fst@algorithm{%
  \coact@algorithm@style
  \let\coact@algorithm@post\@fs@post
  \def\@fs@post{\coact@algorithm@post\vskip-17pt}}
\makeatother
\begin{algorithm}[t]
\caption{COACT-Retarget}
\label{alg:retarget}
\algrenewcommand\algorithmicindent{0.9em}
\begin{algorithmic}[1]
\Require $x^{\mathrm{ref}}_{0:T}$, contact labels $l^{\mathrm{ref,h}}_{0:T}$
\For{each receding-horizon window}
  \While{not converged}
    \State Anneal $\Sigma^{i}$, draw $[U^{i}]_{1:N_W}$, roll out $x^{j}_{0:H}$
           \Comment{\cite{panSPIDERScalablePhysicsInformed2026}}
    \For{$j=1,\ldots,N_W$}
      \State $R^{j}\gets\sum_{t,e}\bigl(r^{\mathrm{coarse}}_{t,e}+r^{\mathrm{fine}}_{t,e}\bigr)$
             \Comment{\eqref{eq:coarse_contact_reward}--\eqref{eq:fine_contact_reward}}
      \State $C^{j}\gets\sum_{t,P\in\{\mathrm{body},\mathrm{leg}\}}w_P\max\bigl(0,\delta_P-\phi_P(x^{j}_t)\bigr)$
      \State $\mathrm{V}^{j}_P\gets\operatorname{Check}_P(x^j_{0:H})$, $\forall P$
      \State $J^{j}\gets J^{j}_{\mathrm{track}}-R^{j}+C^{j}+J_z^j$
             \Comment{\eqref{eq:retarget_cost}}
    \EndFor
    \State $\mathcal{V}\gets\{j:\bigwedge_{P\in\{\mathrm{hand},\mathrm{body},\mathrm{leg}\}}\mathrm{V}^{j}_P\}$
           \Comment{all parts pass}
    \State $\mathcal{E}\gets\lceil\rho N_W\rceil$ lowest $J^{j}$ in $\mathcal{V}$, else least violation
    \State Refit $U^{i+1}$ over $\mathcal{E}$; check early stop
  \EndWhile
  \State Execute leading controls; advance the simulator
\EndFor
\end{algorithmic}
\end{algorithm}
\endgroup

\subsubsection{Multi-Scale Contact Rewards}
\label{sec:multi_scale_contact}
To recover hand--object contact from imperfect kinematic references, we propose multi-scale contact rewards that combine a coarse anchor term with a fine surface term.
The coarse term provides long-range guidance toward the reference contact location, but proximity to this anchor alone does not ensure contact with the object surface.
The fine term uses the object SDF to reward proximity to the surface within a narrow band, refining contact according to the actual object geometry.
Both terms are applied only to end-effectors and frames labeled for contact by $l^{\mathrm{ref,h}}_t$.

For each end-effector $e$, the object-attached anchor and current end-effector contact point in the world frame are,
\vspace{-0.1cm}
{\small
\begin{equation}
  p^{\star,e}_t=p_{o,t}+R(\varrho_{o,t})d^e_t,\qquad
  p^e_t=p_{e,t}+R(\varrho_{e,t})d^e_{\mathrm{loc}}
  \label{eq:contact_points}
\end{equation}
}
where $d^e_{\mathrm{loc}}$ is a fixed end-effector-local contact point and $d^e_t$ expresses its position, computed by reference forward kinematics, in the reference object frame. $R(\cdot)$ maps quaternions to rotation matrices; $(p_{o,t},\varrho_{o,t})$ and $(p_{e,t},\varrho_{e,t})$ are the current object and end-effector poses.
The coarse term combines attraction to the object-attached anchor with palm-orientation alignment,
\vspace{-0.2cm}
{\small
\begin{equation}
  r^{\mathrm{coarse}}_{t,e}=w_c\bigl[(1-\kappa)e^{-\frac{\|p^e_t-p^{\star,e}_t\|}{\sigma_c}}+\kappa\max(0,\hat{n}^e_t\!\cdot\!\hat{u}^e_t)\bigr]l^{\mathrm{ref,h},e}_t
  \label{eq:coarse_contact_reward}
\end{equation}
}
\vspace{-0.1cm}
where $\hat{n}^e_t$ is the palm normal and $\hat{u}^e_t$ is the unit vector from $p^e_t$ toward $p^{\star,e}_t$, so their alignment favors a palm facing the anchor. The weight $w_c$ scales the reward, $\sigma_c$ controls the attraction range, and $\kappa$ balances distance and orientation.

The fine term rewards surface proximity within a narrow SDF band:
\vspace{-0.2cm}
{\small
\begin{equation}
  r^{\mathrm{fine}}_{t,e}=w_f\exp\!\left(-\frac{|\phi_e(x_t)|}{\sigma_f}\right)
  \mathbf{1}\!\left[\phi_e(x_t)\in B\right]l^{\mathrm{ref,h},e}_t
  \label{eq:fine_contact_reward}
\end{equation}
}
\vspace{-0.1cm}
where $\sigma_f$ controls the surface-band scale and $w_f$ is its reward weight.
The asymmetric band $B=[-1,+3]\,\mathrm{mm}$ maintains a continuous, distance-sensitive reward across the contact interface, keeping contact guidance active when the hand engages the surface under rigid-body contact dynamics.
We additionally decay the fine reward over the final $15\%$ of the trajectory to encourage a smooth release.

\subsubsection{Part-Wise Penetration Constraints}
\label{sec:partwise_constraints}
Contact rewards encourage hand--object proximity but do not impose explicit penetration limits.
We therefore partition the collision geometry into $P\in\{\mathrm{hand},\mathrm{body},\mathrm{leg}\}$, with different tolerances for intended hand contacts and non-contact body parts.
For $P\in\{\mathrm{body},\mathrm{leg}\}$, the soft penalty $c^{\mathrm{pen}}_{t,P}=\max(0,\delta_P-\phi_P(x_t))$ provides a graded cost when clearance falls below the activation margin $\delta_P$.
Hands are excluded to avoid counteracting the contact reward.
During candidate selection, the hard filter in Alg.~\ref{alg:retarget} enforces a per-group SDF threshold, a maximum fraction of rollout frames below it, and a hard SDF floor, with settings given in Sec.~\ref{sec:implementation}.
Here, $\operatorname{Check}_P(x^j_{0:H})$ returns $1$ if group $P$'s SDF never falls below its hard floor during the rollout and the fraction of frames below its SDF threshold does not exceed the allowed rate, and $0$ otherwise.
The legs require positive clearance with a small violation allowance, the body allows no threshold violations, and the hands permit shallow penetration.
The optimized reference trajectory $\tilde{x}_{0:T}$ provides improved hand--object contact with reduced penetration for downstream policy learning.

\vspace{-0.2cm}
\subsection{COACT-Policy: Closed-Loop Collaborative Policy}
\label{sec:closed_loop_learning}

While COACT-Retarget provides dynamically feasible
reference humanoid motions, executing collaborative
carrying with a free rigid object requires
the humanoid to interact with the object at appropriate times, share the load, and adapt to partner-induced changes.
To address this challenge, COACT-Policy couples COACT-Partner, a physical co-actuating partner, with the policy through shared object dynamics.
Demonstrated contact labels gate partner assistance ($l^{\mathrm{ref,p}}_t$) and supervise humanoid contact behavior ($l^{\mathrm{ref,h}}_t$), synchronizing both throughout lifting, carrying, and placement.

\subsubsection{COACT-Partner: Physical Co-Actuating Partner}
\label{sec:parametric_partner}

COACT-Partner is a force-centric controller that reacts to the object's live state and applies a bounded force and torque directly to it.
Unlike COLA’s PD-controlled support body~\cite{duLearningHumanHumanoidCoordination2025}, COACT-Partner applies contact-guided, rate-limited forces at an object-local anchor with an explicit gravity-compensation share.
The controller acts at an object-local anchor $a$ obtained offline as the clip median of the mean demonstrated partner-hand position in the object frame over partner-contact frames.
Let $(p_a,v_a)$ and $(p^{\star}_a,v^{\star}_a)$ denote the live and reference anchor states, obtained by transporting $a$ with the simulated object state and reference object trajectory of $\tilde{x}_{0:T}$, and let $\Delta p_a=p^{\star}_a-p_a$ and $\Delta v_a=v^{\star}_a-v_a$ denote the tracking errors.
A PD controller produces the force as a critically damped tracking correction plus a nominal gravity-sharing term:
\vspace{-0.1cm}
\begin{equation}
\begin{aligned}
F_{\mathrm{corr}}&=m\left(\omega_n^2\Delta p_a+2\zeta\omega_n\Delta v_a\right),\\
F_{\mathrm{app}}&=\mathcal{R}\!\left\{\operatorname{clip}_{F_{\max}}\!\left(\operatorname{clip}_{c_{\max}}\!\left(F_{\mathrm{corr}}\right)+\alpha\, m g\hat{z}\right)\right\},
\end{aligned}
\label{eq:partner_command}
\end{equation}
where $m$ is the live randomized object mass, $g$ is gravitational acceleration, $\hat{z}$ is the vertical unit vector, $\zeta=1$ gives critical damping at natural frequency $\omega_n$, $\alpha$ sets the nominal gravity-share fraction, $\operatorname{clip}$ bounds the vector norm with correction-force cap $c_{\max}$ and total-force cap $F_{\max}$, and $\mathcal{R}$ enforces a force-rate limit $\dot{F}_{\max}$ and activation ramp $t_{\mathrm{ramp}}$ gated by $l^{\mathrm{ref,p}}_t$.
Because $F_{\mathrm{app}}$ acts at the live anchor $p_a$ rather than the object center of mass, it naturally induces a torque $\boldsymbol{\tau}=r_{\mathrm{live}}\times F_{\mathrm{app}}$ about the center of mass with lever arm $r_{\mathrm{live}}$, which is norm-clipped to $\tau_{\max}$ without an independent rotational controller.
The policy does not observe these forces directly but adapts to the partner’s actions through their effects on the object.

\subsubsection{Contact-Guided Coordination Training}
\label{sec:coordination_training}

We train a \emph{single} policy across all objects and clips, conditioned on each clip only through its observations: the current-frame reference robot state and contact label, measured object pose, and proprioception.
At each step, the policy emits scaled joint-position targets and is optimized with the reward
\vspace{-0.2cm}
\begin{equation}
r=r_{\mathrm{track}}+r_{\mathrm{contact}}+r_{\mathrm{reg}}.
  \label{eq:policy_reward}
\end{equation}
\vspace{-0.2cm}

The tracking term $r_{\mathrm{track}}$ scores exponential error kernels against $\tilde{x}_{0:T}$, while $r_{\mathrm{reg}}$ penalizes non-smooth control and undesired contacts~\cite{wuSUGARScalableHumanvideodriven2026a}.
The contact term $r_{\mathrm{contact}}=w_h\mathbf{1}[\hat l_t^{\mathrm{h}}=l_t^{\mathrm{ref,h}}]$ rewards label agreement, where $\hat l_t^{\mathrm{h}}=1$ when each hand's peak object-contact force magnitude over the last three sensor samples exceeds $F_{\mathrm{th}}$.

The policy responds to partner-induced changes using object-state observations and proprioception, while its actions affect the object through contact.
With bounded partner assistance, the object-tracking and contact rewards require the humanoid to co-actuate the free object, share its load, and realize demonstrated contact transitions.
Randomized object mass and friction, together with intermittent pushes to the humanoid and object during contact, train the policy to adapt to interaction dynamics.

\subsubsection{Teacher--Student Distillation and Deployment}
\label{sec:teacher_student}

We first optimize a privileged teacher policy $\pi_{\mathrm{teach}}$ with proximal policy optimization (PPO), using privileged observations that include future reference targets and simulator states (live object and body velocities, base linear velocity, and full-body poses).
Following~\cite{wuSUGARScalableHumanvideodriven2026a}, we distill the teacher into a deployable student policy $\pi_{\mathrm{stud}}$ via behavior cloning, transitioning to full PPO after $2000$ updates.
The student observes proprioception, the current-frame reference state, and the object pose.
At deployment, the student tracks $\tilde{x}_{0:T}$ through all three phases without privileged observations or partner-force measurements.

\vspace{-0.1cm}
\subsection{Implementation Details}
\label{sec:implementation}

Closed-loop policy training and simulation evaluation run at $50$~Hz in Isaac Lab with a 29-DoF Unitree G1 and a free rigid-body object.
Both dynamic retargeting and policy learning use $50$ CORE4D-Real human--human collaborative-carrying sequences spanning $11$ object types across boxes, buckets, desks, and chairs.
To construct a whole-hand collaborative carrying dataset for the G1 embodiment, we select two-person carrying sequences stratified by object type, excluding recordings with object-pose drift or single-hand interactions.

COACT-Retarget uses $H=0.8$~s windows with $0.1$~s knot spacing, $N_W=1024$, $\rho=0.1$, and at most $32$ iterations.
Contact settings are $(w_c,\sigma_c,\kappa)=(5,0.30\,\mathrm{m},0.30)$ and $(w_f,\sigma_f)=(1.5,1.5\,\mathrm{mm})$, with $(w_P,\delta_P)=(2,20\,\mathrm{mm})$ for body and leg clearance.
The filter tuples (SDF threshold [mm], allowed violation rate, hard SDF floor [mm]) are $(-5,0\%,-5)$, $(+5,2\%,-5)$, and $(-10,10\%,-20)$ for the body, legs, and hands, respectively.

For COACT-Policy learning, we use a 3-layer MLP with hidden dimensions of $512$, $256$, and $128$.
The teacher and student train for $20$k iterations on four and two NVIDIA RTX 4090 GPUs, respectively, with $4{,}096$ environments per GPU and $24$ steps per environment per update.
PPO uses an initial learning rate of $10^{-3}$, discount factor $\gamma=0.99$, generalized advantage estimation parameter $\lambda=0.95$, and policy-ratio clipping of $0.2$.
The partner uses $\omega_n=4$~rad/s, $\alpha \sim \mathcal{U}[0.4, 0.6]$ drawn at each reset, $c_{\max}=15$~N, $F_{\max}=75$~N, $\dot F_{\max}=120$~N/s, and $\tau_{\max}=6$~N\,m.
The contact reward uses $w_h=1$ and $F_{\mathrm{th}}=0.1$~N per hand.
During training, we randomize material properties at reset and apply external velocity perturbations to the robot and object.

\vspace{-0.1cm}
\section{EXPERIMENTS}

\subsection{Experimental Setup}

\noindent\textbf{Evaluation data.}
Simulation evaluations use the $50$ CORE4D-G1 training references and an augmented test set of $12$ references with initial object pose perturbations of $0.2$~m in translation and $\pm45^\circ$ in yaw.
We synthesize valid trajectories by smoothly interpolating from each perturbed initial object pose to the demonstrated contact endpoint. All methods are evaluated on the same resulting cases.

\noindent\textbf{Baselines and protocol.}
We compare COACT-Retarget against OmniRetarget~\cite{yangOmniRetargetInteractionPreservingData2025},
DynaRetarget~\cite{dhedinDynaRetargetDynamicallyFeasibleRetargeting2026}, and
GMR~\cite{araujoRetargetingMattersGeneral2025}.
For the downstream comparison, each method's reference is passed to the same COACT-Policy learning setup with identical partner model, reward, and training budget.
For COACT-Policy, we compare adapted partner models and ablate contact guidance and teacher supervision.
COLA-Partner follows COLA's horizontal integral-control design~\cite{duLearningHumanHumanoidCoordination2025}, adding integral action to our position PD controller in place of velocity-command tracking.
PAINT-Partner follows PAINT's external-wrench formulation~\cite{caoPAINTPartneragnosticIntentaware2026}, using reference-derived feedforward forces for the demonstrated motions.
The \emph{w/o partner} variant disables assistance during both training and evaluation.
All simulation policy results report the student policy with the partner enabled unless stated otherwise.
Each trajectory is evaluated and averaged across 3 random seeds with $512$ environments per seed.

\begin{figure}[t]
  \centering
  \includegraphics[width=\linewidth]{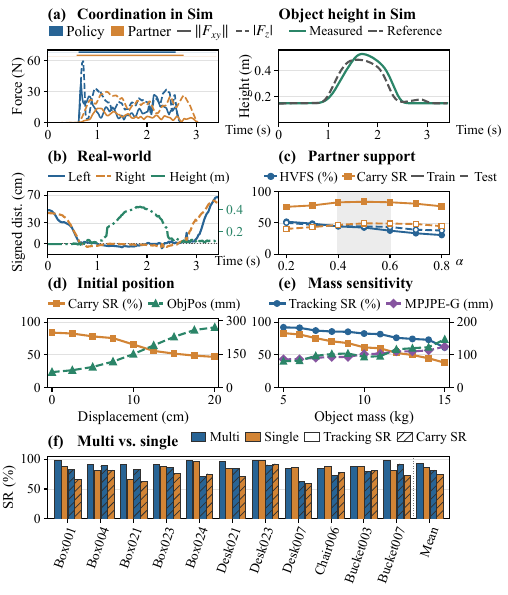}
  \vspace{-0.9cm}
  \caption{Policy evaluation. (a) Simulated interaction forces and object height.
  (b) Real-world hand--object signed distance and object height.
  (c) HVFS and Carry SR versus the partner's gravity-share fraction $\alpha$ with a fixed
  policy; solid/dashed: training references/the augmented test set, shading: training range.
  (d) Initial-position sensitivity at $5$~kg.
  (e) Mass sensitivity; Carry SR/ObjPos as in (d).
  (f) Multi-/single-object success rates over $11$ objects.}
  \label{fig:policy_evaluation}
\vspace{-0.8cm}
\end{figure}

\vspace{-0.05cm}
\noindent\textbf{Metrics.}
In simulation, the \emph{motion tracking success rate} (Tracking SR) requires every frame to stay within $0.3$~m on torso, wrist/ankle, and object positions and within $0.8$ on torso tilt and object orientation.
The \emph{collaborative carrying success rate} (Carry SR) requires final placement error $\leq0.10$~m, mean CoM lift during transport $>50\%$ of the reference lift, and nonzero humanoid hand--object contact force during placing while the object bottom is $>3$~cm above ground. 
Carry SR also requires stable placement over a time window (object bottom clearance $\leq1$~cm, speed $\leq3$~cm/s, and position drift $\leq5$~mm) and sustained release of both hands with zero contact forces.
MPJPE-G is the global whole-body mean per-joint position error, and ObjPos is the final object-to-target Euclidean distance.
The \emph{humanoid vertical force share} (HVFS) is $100F_z^{\mathrm{h}}/(F_z^{\mathrm{h}}+F_z^{\mathrm{p}})$, where $F_z^{\mathrm{h}}$ and $F_z^{\mathrm{p}}$ are the vertical forces applied to the object by the humanoid and partner, respectively.
For retargeting, contact fidelity is the fraction of reference contact frames with detected hand--object contact and penetration of at most $10$~mm.
Penetration fraction is the fraction of all frames with hand--object penetration deeper than $10$~mm.
We also report the mean per-sequence maximum penetration depth.

\label{sec:main_result}

\vspace{-0.1cm}
\subsection{Evaluation of COACT-Policy}

\vspace{-0.1cm}
\noindent\textbf{Complete collaborative carrying.}
COACT completes the full lifting--carrying--placing sequence from an initially unheld object, reaching $83.86\%$ Carry SR on the training references and $46.78\%$ on the augmented test set (Table~\ref{tab:partner_comparison}).
In the rollout of \figpanelref{fig:policy_evaluation}{(a)}, the humanoid's force appears at contact and vanishes after release, overlapping the partner's assistance, while the object height follows the reference.

\noindent\textbf{Closed-loop partner coupling.}
Partner coupling is essential for the complete task (Table~\ref{tab:partner_comparison}).
Without it, the policy still meets the tracking criteria on most training references but loses about a third of its successful carries, and it almost always fails under perturbation.
COACT-Partner outperforms both adapted partners on all success and error metrics while leaving the humanoid a similar vertical share, so its advantage comes from how assistance is applied rather than from how much load the partner takes.

\begin{table}[t]
  \vspace{0pt}
  \centering
  \caption{Partner-model comparison. Upper: training set. Lower ($^{\dagger}$): augmented test set. 
  Partner modules are adapted to COACT. For~the \textnormal{\itshape w/o partner} rows, HVFS is 100\% by definition.}
  \vspace{-0.3cm}
  \label{tab:partner_comparison}
  \fontsize{7.4}{7.8}\selectfont
  \newcommand{\compactcellfont}{\fontsize{7.1}{7.8}\selectfont}
  \renewcommand{\arraystretch}{0.90}
  \setlength{\tabcolsep}{1.5pt}
  \setlength{\aboverulesep}{0.8pt}
  \setlength{\belowrulesep}{0.8pt}
  \begin{tabular*}{\columnwidth}{@{\extracolsep{\fill}}>{\compactcellfont}l*{5}{>{\compactcellfont}r}@{\hspace{3pt}}}
    \toprule
    Method & \makecell[c]{Tracking SR\\(\%) $\uparrow$} & \makecell[c]{Carry SR\\(\%) $\uparrow$} & \makecell[c]{MPJPE-G\\(mm) $\downarrow$} & \makecell[c]{ObjPos\\(mm) $\downarrow$} & \makecell[c]{HVFS\\(\%)} \\
    \midrule
    w/o partner & 80.58 & 56.37 & 127.40 & 146.04 & 100.00\rlap{\textsuperscript{*}} \\
    COLA-Partner & 85.15 & 73.52 & 91.57 & 87.22 & 40.46 \\
    PAINT-Partner & 82.55 & 73.85 & 96.47 & 99.13 & 41.43 \\
    \textbf{COACT (Ours)} & \textbf{91.81} & \textbf{83.86} & \textbf{82.63} & \textbf{76.49} & 41.89 \\
    \midrule
    w/o partner$^{\dagger}$ & 8.33 & 6.01 & 316.36 & 313.03 & 100.00\rlap{\textsuperscript{*}} \\
    COLA-Partner$^{\dagger}$ & 49.03 & 37.29 & 146.83 & 179.73 & 39.96 \\
    PAINT-Partner$^{\dagger}$ & 35.40 & 28.48 & 170.66 & 166.74 & 42.50 \\
    \textbf{COACT (Ours)}$^{\dagger}$ & \textbf{59.93} & \textbf{46.78} & \textbf{117.93} & \textbf{103.07} & 44.41 \\
    \bottomrule
  \end{tabular*}
\vspace{-0.7cm}
\end{table}

\noindent\textbf{Adaptive load sharing.}
The humanoid bears about two-fifths of the vertical load in both settings (Table~\ref{tab:partner_comparison}) and adapts this share to the partner.
When we sweep the partner's gravity-share fraction $\alpha$ of a frozen policy from $0.2$ to $0.8$, beyond the training range $[0.4,0.6]$, HVFS decreases monotonically from about half to $30$--$38\%$, while Carry SR stays within ten points of its peak (\figpanelref{fig:policy_evaluation}{(c)}).
Since the policy never observes partner forces, this adaptation emerges from the physical coupling through the shared object.

\noindent\textbf{Robustness to initial pose and payload.}
We stress-test the policy beyond its training references (\figpanelref{fig:policy_evaluation}{(d),~(e)}).
When a $5$~kg object is translated forward, backward, left, or right by up to $0.2$~m, with references regenerated as for the augmented test set, Carry SR degrades gracefully, staying $78.07\%$ within $5$~cm.
For payloads, Carry SR of a policy trained on $5$--$10$~kg declines at a similar rate inside and beyond this range, with no sharp drop at $10$~kg, and still reaches $37.9\%$ at $15$~kg, $50\%$ above the heaviest training mass.

\noindent\textbf{One policy across objects.}
Compared with $11$ object-specific specialists trained with the same per-stage iteration budget, our single multi-object policy raises mean Carry SR by $7.22$ points and outperforms the specialists on most objects (\figpanelref{fig:policy_evaluation}{(f)}), suggesting that contact and load-sharing behaviors transfer across objects instead of being diluted.

\vspace{-0.2cm}
\subsection{Evaluation of COACT-Retarget}

\begin{figure}[!t]
  \centering
  \includegraphics[width=\linewidth]{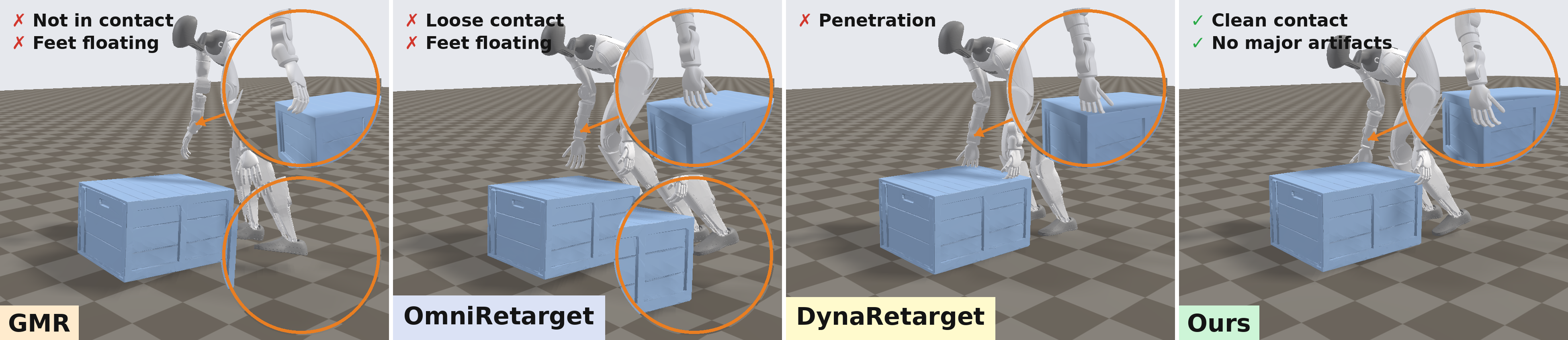}
  \vspace{-0.7cm}
  \caption{Qualitative retargeting comparison on the same representative lifting frame from CORE4D-G1. Orange circular insets highlight hand--object and foot--ground regions, and annotations identify contact and physical artifacts.}
  \label{fig:retarget_qualitative}
\vspace{-0.4cm}
\end{figure}

\noindent\textbf{Reference quality.}
COACT-Retarget is the only method that combines high contact fidelity with low penetration (Table~\ref{tab:retarget_policy}).
Kinematic retargeting leaves contact unreliable and penetration frequent, with GMR almost never producing valid contact.
COACT-Retarget also exceeds the contact fidelity of the dynamics-based DynaRetarget by $15$ points and cuts OmniRetarget's penetration fraction by more than an order of magnitude, which we attribute to the multi-scale contact rewards and part-wise penetration constraints.
Fig.~\ref{fig:retarget_qualitative} shows the same pattern at the frame level. The kinematic baselines leave the hand off the surface with the feet floating, DynaRetarget penetrates the object, and only COACT-Retarget achieves clean contact with planted feet.

\noindent\textbf{Downstream policy learning.}
Because only the reference source varies, the differences in the policy blocks of Table~\ref{tab:retarget_policy} are attributable to the references themselves.
COACT-Retarget references yield the best success and error metrics on both evaluation sets, and their margin over the strongest baseline, OmniRetarget, widens under perturbation in both Carry SR and tracking and placement errors.
Contact-guided, dynamically feasible references thus matter most when the policy must deviate from the demonstration.

\begin{table}[t]
  \vspace{6pt}
  \centering
  \caption{Retargeting and downstream policy comparison. Only the reference source varies; all other settings are held fixed.}
  \vspace{-0.3cm}
  \label{tab:retarget_policy}
  \fontsize{8}{8.4}\selectfont
  \newcommand{\compactcellfont}{\fontsize{7.5}{8.4}\selectfont}
  \renewcommand{\arraystretch}{1.0}
  \setlength{\tabcolsep}{1.5pt}
  \setlength{\aboverulesep}{1.3pt}
  \setlength{\belowrulesep}{1.3pt}
  \begin{tabular*}{\columnwidth}{@{\extracolsep{\fill}}>{\compactcellfont}l>{\compactcellfont}r>{\compactcellfont}r>{\compactcellfont}r>{\compactcellfont}r@{}}
    \toprule
    Metric & COACT-Retarget & OmniRetarget & DynaRetarget & GMR \\
    \midrule
    \multicolumn{5}{l}{\compactcellfont\textit{Retargeting quality}} \\
    Contact (\%) $\uparrow$ & \textbf{80.49} & 49.09 & 65.09 & 1.33 \\
    Penetration (\%) $\downarrow$ & \textbf{1.74} & 25.68 & 13.36 & 33.44 \\
    Max pen. (mm) $\downarrow$ & \textbf{12.01} & 21.61 & 25.36 & 47.37 \\
    \midrule
    \multicolumn{5}{l}{\compactcellfont\textit{Policy: training references}} \\
    Tracking SR (\%) $\uparrow$ & \textbf{91.81} & 84.87 & 77.80 & 59.53 \\
    Carry SR (\%) $\uparrow$ & \textbf{83.86} & 75.38 & 60.63 & 20.65 \\
    MPJPE-G (mm) $\downarrow$ & \textbf{82.63} & 107.07 & 115.48 & 156.57 \\
    ObjPos (mm) $\downarrow$ & \textbf{76.49} & 102.66 & 121.96 & 159.41 \\
    HVFS (\%) & 41.89 & 40.87 & 20.50 & 36.73 \\
    \midrule
    \multicolumn{5}{l}{\compactcellfont\textit{Policy: augmented test set}} \\
    Tracking SR (\%) $\uparrow$ & \textbf{59.93} & 49.98 & 52.18 & 42.19 \\
    Carry SR (\%) $\uparrow$ & \textbf{46.78} & 36.25 & 17.64 & 19.93 \\
    MPJPE-G (mm) $\downarrow$ & \textbf{117.93} & 188.27 & 247.67 & 237.64 \\
    ObjPos (mm) $\downarrow$ & \textbf{103.07} & 154.93 & 241.02 & 244.31 \\
    HVFS (\%) & 44.41 & 38.14 & 28.37 & 38.45 \\ 
    \bottomrule
  \end{tabular*}
\vspace{-0.6cm}
\end{table}

\vspace{-0.15cm}
\subsection{Ablation Studies}

\vspace{-0.1cm}
\noindent\textbf{Contact guidance in policy learning.}
Removing either contact label causes a distinct failure (Table~\ref{tab:ablation_r021}).
Without the humanoid contact reward ($l^{\mathrm{ref,h}}_t$), tracking error is nearly unchanged, yet HVFS drops by about a quarter and Carry SR falls, confirming that a pose reference alone does not specify the interaction force.
Without the partner gate ($l^{\mathrm{ref,p}}_t$), the partner also acts outside the demonstrated contact phases, which lowers Carry SR by a similar margin.

\begin{table}[t]
  \centering
  \caption{Policy ablations on training references. Removing $l^{\mathrm{ref,p}}_t$ disables partner gating; removing $l^{\mathrm{ref,h}}_t$ removes the humanoid contact reward.}
  \vspace{-0.3cm}
  \label{tab:ablation_r021}
  \fontsize{8}{8.4}\selectfont
  \newcommand{\compactcellfont}{\fontsize{7.5}{8.4}\selectfont}
  \renewcommand{\arraystretch}{1.0}
  \setlength{\tabcolsep}{1.5pt}
  \setlength{\aboverulesep}{1.3pt}
  \setlength{\belowrulesep}{1.3pt}
  \begin{tabular*}{\columnwidth}{@{\extracolsep{\fill}}>{\compactcellfont}l*{5}{>{\compactcellfont}r}@{}}
    \toprule
    Method & \makecell[c]{Tracking SR\\(\%) $\uparrow$} & \makecell[c]{Carry SR\\(\%) $\uparrow$} & \makecell[c]{MPJPE-G\\(mm) $\downarrow$} & \makecell[c]{ObjPos\\(mm) $\downarrow$} & \makecell[c]{HVFS\\(\%)} \\
    \midrule
    w/o $l^{\mathrm{ref,p}}_t$ & 86.30 & 76.31 & 93.08 & 87.51 & 40.72 \\
    w/o $l^{\mathrm{ref,h}}_t$ & 82.00 & 75.13 & 83.90 & 93.86 & 31.72 \\
    w/o Teacher & 73.68 & 64.43 & 137.05 & 117.04 & 34.32 \\
    \textbf{COACT (Ours)} & \textbf{91.81} & \textbf{83.86} & \textbf{82.63} & \textbf{76.49} & 41.89 \\
    \bottomrule
  \end{tabular*}
\vspace{-0.3cm}
\end{table}

\noindent\textbf{Teacher supervision.}
Without teacher supervision, both tracking and carrying fall markedly (Table~\ref{tab:ablation_r021}), confirming that privileged supervision lets the student acquire coordinated whole-body and object control.

\noindent\textbf{Multi-Scale Contact Reward.}
Removing either contact scale costs about a third of contact fidelity (Table~\ref{tab:ablation50}).
Without the coarse term, whole-body MPJPE more than triples, consistent with losing long-range guidance toward the contact location. 
Removing the fine term, the hands reach the contact region but often fail to settle on the surface.
The coarse term thus brings the hand to the contact region, and the
fine term closes the gap to the object geometry.

\noindent\textbf{Part-Wise Penetration Constraints.}
Both constraint layers matter.
Removing either the hard filter or the soft penalty increases penetration with contact fidelity nearly intact, and also raises whole-body MPJPE by roughly 30–50\% (Table~\ref{tab:ablation50}), as unintended body–object collisions in the dynamic rollouts push the robot off its reference.

\begin{table}[t]
\centering
\caption{Leave-one-out ablation of COACT-Retarget on CORE4D-G1. \textbf{Best} in bold, \underline{second best} underlined.}
\vspace{-0.3cm}
\label{tab:ablation50}
\scriptsize
\setlength{\tabcolsep}{3pt}
\begin{tabular}{l ccccc}
\toprule
Metric & COACT-Retarget & w/o fine & w/o coarse & w/o hard & w/o soft \\
\midrule
Contact (\%) $\uparrow$ & \textbf{80.49} & 53.96 & 50.78 & 78.18 & \underline{79.23} \\
Penetration (\%) $\downarrow$ & \textbf{1.74} & 2.83 & \underline{1.81} & 2.00 & 2.19 \\
Max pen.\ (mm) $\downarrow$ & \underline{12.01} & 16.05 & \textbf{11.63} & 13.13 & 12.82 \\
\midrule
MPJPE (cm) $\downarrow$ & \underline{12.34} & \textbf{12.28} & 40.07 & 18.88 & 15.78 \\
\midrule
Obj pos (cm) $\downarrow$ & \underline{10.24} & \textbf{9.74} & 10.34 & 10.74 & 10.48 \\
Obj ori (deg) $\downarrow$ & 5.01 & \underline{4.99} & \textbf{4.97} & 5.47 & 5.60 \\
\bottomrule
\end{tabular}
\vspace{-0.6cm}
\end{table}

\vspace{-0.2cm}
\subsection{Real-World Experiments}
\label{sec:real_world}

We deploy the same policy on a Unitree G1 at $50$~Hz, with VIVE Trackers providing robot pelvis and object poses. Three subjects acted as human partners, each executing the partner role of the demonstrated carrying motion that the humanoid tracks.
A trial succeeds only if the humanoid starts without robot--object contact and completes the whole lifting-carrying-placing task.
Trained only with the simulated partner, the policy transfers directly to human partners and succeeds in $76.67\%$ of the $120$ trials (Table~\ref{tab:real_world}, \figpanelref{fig:real_world}{(a),~(b)}).
\figpanelref{fig:policy_evaluation}{(b)} confirms that the policy autonomously establishes, maintains, and releases contact on hardware. The hand--object distance closes to near zero before lifting, stays there through transport, and grows only after placement.
The same policy also handles shifted and rotated initial poses and payloads from $1$ to $10$~kg (\figpanelref{fig:real_world}{(c),~(d)}).

\begin{table}[t]
  \centering
  \caption{Single-policy real-world collaborative carrying.
  A--C: successes/trials; ---: no sequence. SR: pooled success rate.}
  \vspace{-0.3cm}
  \label{tab:real_world}
  \footnotesize
  \setlength{\tabcolsep}{1.5pt}
  \renewcommand{\arraystretch}{0.95}
  \begin{tabular*}{\columnwidth}{@{\extracolsep{\fill}}l*{7}{c}@{}}
    \toprule
    Seq. & Box001 & Box021 & Box024 & Box023 & Bucket003 & Desk021 & Desk023 \\
    \midrule
    A & 8/10 & 8/10 & 6/10 & 7/10 & 7/10 & 8/10 & 7/10 \\
    B & 8/10 & 9/10 & 8/10 & ---  & ---  & ---  & ---  \\
    C & 9/10 & 7/10 & ---  & ---  & ---  & ---  & ---  \\
    \midrule
    SR (\%) & 83.33 & 80.00 & 70.00 & 70.00 & 70.00 & 80.00 & 70.00 \\
    \midrule
    \multicolumn{8}{@{}l@{}}{\textbf{Overall (12 sequences): 92/120 (76.67\%)}} \\
    \bottomrule
  \end{tabular*}
  \vspace{-0.7cm}
\end{table}

\vspace{-0.10cm}
\section{CONCLUSIONS}
This paper advances human--humanoid collaborative carrying with a single humanoid policy that completes lifting, carrying, and placing while autonomously establishing, maintaining, and releasing contact.
Our key contribution is COACT, a contact-guided motion tracking framework that learns these behaviors from human demonstrations through two synergistic components: COACT-Retarget and COACT-Policy.
Experiments across various objects validate COACT with a single policy, achieving an 83.86\% success rate on the simulator and 76.67\% on the real-world Unitree G1.

Our current setting derives references from demonstrations and measures object pose with external tracking. Extending COACT to onboard perception would remove this dependency and carry the same contact-guided formulation into unstructured environments.
Beyond demonstrated motions, future work could integrate intent inference and high-level planning to collaborate with evolving human goals.
We hope this work inspires further research on physical human--humanoid interaction across broader tasks and environments.

\vspace{-0.2cm}
\bibliographystyle{IEEEtran}
\begingroup
\linespread{0.94}
\bibliography{main}
\endgroup

\end{document}